\documentclass[11pt,letterpaper]{article}
\pdfoutput=1 
\usepackage{naaclhlt2016}
\usepackage{times}
\usepackage{latexsym}

\usepackage{booktabs}
\usepackage{tabularx}

\naaclfinalcopy 
\def\naaclpaperid{***} 

\title{Sentence Similarity Measures for Fine-Grained Estimation of Topical Relevance in Learner Essays}

\author{Marek Rei\\
	    The ALTA Institute\\
	    Computer Laboratory\\
	    University of Cambridge\\
        United Kingdom\\
	    {\tt marek.rei@cl.cam.ac.uk}
	  \And
	Ronan Cummins\\
	    The ALTA Institute\\
	    Computer Laboratory\\
	    University of Cambridge\\
        United Kingdom\\
  {\tt ronan.cummins@cl.cam.ac.uk}}

\date{}

\begin{document}

\maketitle

\begin{abstract}
We investigate the task of assessing sentence-level prompt relevance in learner essays.
Various systems using word overlap, neural embeddings and neural compositional models are evaluated on two datasets of learner writing.
We propose a new method for sentence-level similarity calculation, which learns to adjust the weights of pre-trained word embeddings for a specific task, achieving substantially higher accuracy compared to other relevant baselines.

\end{abstract}

\section{Introduction}

Evaluating the relevance of learner essays with respect to the assigned prompt is an important part of automated writing assessment \cite{Higgins2006,Briscoe2010}. Students with limited relevant vocabulary may attempt to shift the topic of the essay in a more familiar direction, which grammatical error detection systems are not able to capture. In an automated examination framework, this weakness could be further exploited by memorising a grammatically correct essay and presenting it in response to \emph{any} prompt. Being able to detect topical relevance can help prevent such weaknesses, provide useful feedback to the students, and is also a step towards evaluating more creative aspects of learner writing.

Most existing work on assigning topical relevance scores has been done using supervised methods. 
\newcite{Persing2014} trained a linear regression model for detecting relevance to each prompt, but this approach requires substantial training data for all the possible prompts. 
\newcite{Higgins2006} addressed off-topic detection by measuring the cosine similarity between tf-idf vector representations of the prompt and the entire essay. 
However, as this method only captures similarity using exact matching at the word-level, it can miss many topically relevant word occurrences in the essay. 
In order to overcome this limitation, \newcite{Louis2010} investigated a number of methods that expand the prompt with related words, such as morphological variations. 
Ideally, the assessment system should be able to handle the introduction of new prompts, i.e. ones for which no previous data exists. 
This allows the list of available topics to be edited dynamically, and students or teachers can insert their own unique prompts for every essay. We can achieve this by constructing an unsupervised function that measures similarity between the prompt and the learner writing.

While previous work on prompt relevance assessment has mostly focussed on full essays, scoring individual sentences for prompt relevance has been relatively underexplored.
\newcite{Higgins2004} used a supervised SVM classifier to train a binary sentence-based relevance model with 18 sentence-level features. 
We extend this line of work and investigate unsupervised methods using neural embeddings for the task of assessing topical relevance of individual sentences.
By providing sentence-level feedback, our approach is able to highlight specific areas of the text that require more attention, as opposed to showing a single overall score. Sentence-based relevance scores could also be used for estimating coherence in an essay, or be combined with a more general score for indicating sentence quality \cite{Andersen2013}.

In the following sections we explore a number of alternative similarity functions for this task. The evaluation of the methods was performed on two different publicly available datasets and revealed that alternative approaches are required, depending on the nature of the prompts. We propose a new method which achieves substantially better performance on one of the datasets, and construct a combination approach which provides more robust results independent of the prompt type.

\section{Relevance Scoring Methods}

The systems receive the prompt and a single sentence as input, and aim to provide a score representing the topical relevance of the sentence, with a higher value corresponding to more confidence in the sentence being relevant. For most of the following methods, both the sentence and the prompt are mapped into vector representations and cosine is used to measure their similarity.

\subsection{Baseline methods} 
The simplest baseline we use is a random system where the score between each sentence and prompt is randomly assigned.
In addition, we evaluate the majority class baseline, where the highest score is always assigned to the prompt 
in the dataset which has most sentences associated with it. It is important that any engineered system surpasses the performance of these trivial baselines.

\subsection{TF-IDF}
\label{sec:tfidf}

TF-IDF \cite{SparckJones1972} is a well-established method of constructing document vectors for information retrieval. It assigns the weight of each word to be the multiplication of its term frequency and inverse document frequency (IDF). We adapt IDF for sentence similarity by using the following formula:

$$
IDF(w) = \log(\frac{N}{1 + n_w})
$$

\noindent where $N$ is the total number of sentences in a corpus and $n_w$ is the number of sentences where the target word $w$ occurs. Intuitively, this will assign low weights to very frequent words, such as determiners and prepositions, and assign higher weights to rare words. In order to obtain reliable sentence-level frequency counts, we use the British National Corpus (BNC, \newcite{Burnard2007a}) which contains 100 million words of English from various sources.

\subsection{Word2Vec}
\label{sec:word2vec}

Word2Vec \cite{Mikolov2013a} is a useful tool for efficiently learning distributed vector representations of words from a large corpus of plain text. We make use of the CBOW variant, which maps each word to a vector space and uses the vectors of the surrounding words to predict the target word. This results in words frequently occurring in similar contexts also having more similar vectors. 
To create a vector for a sentence or a document, each word in the document is mapped to a corresponding vector, and these vectors are then summed together.

While the TF-IDF vectors are sparse and essentially measure a weighted word overlap between the prompt and the sentence, Word2Vec vectors are able to capture the semantics of similar words without requiring perfect matches. 
In the experiments we use the pretrained vectors that are publicly available, trained on 100 billion words of news text, and containing 300-dimensional vectors for 3 million unique words and phrases.\footnote{https://code.google.com/archive/p/word2vec/}

\subsection{IDF-Embeddings}

We experiment with combining the benefits of both Word2Vec and TF-IDF. While Word2Vec vectors are better at capturing the generalised meaning of each word, summing them together assigns equal weight to all words. This is not ideal for our task -- for example, function words will likely have a lower impact on prompt relevance, compared to more specific rare words.

We hypothesise that weighting all word vectors individually during the addition can better reflect the contribution of specific words. To achieve this, we scale each word vector by the corresponding IDF weight for that word, following the formula in Section \ref{sec:tfidf}. This will still map the sentence to a distributed semantic vector, but more frequent words have a lower impact on the result.

\subsection{Skip-Thoughts}

Skip-Thoughts \cite{Kiros} is a more advanced neural network model for learning distributed sentence representations.
A single sentence is first mapped to a vector by applying a Gated Recurrent Unit \cite{Cho2014a}, which learns a composition function for mapping individual word embeddings to a single sentence representation. The resulting vector is used as input to a decoder which tries to predict words in the previous and the next sentence.
The model is trained as a single network, and the GRU encoder learns to map each sentence to a vector that is useful for predicting the content of surrounding sentences.

We make use of the publicly available pretrained model\footnote{https://github.com/ryankiros/skip-thoughts} for generating sentence vectors, which is trained on 985 million words of unpublished literature from the BookCorpus \cite{Zhu2015}.

\subsection{Weighted-Embeddings}

We now propose a new method for constructing vector representations, based on insights from all the previous methods.
IDF-Embeddings already introduced the idea that words should have different weights when summing them for a sentence representation. 
Instead of using the heuristic IDF formula, we suggest learning these weights automatically in a data-driven fashion. 

Each word is assigned a separate weight, initially set to $1$, which is used for scaling its vector.
Next, we construct an unsupervised learning framework for gradually adjusting these weights for all words.
The task we use is inspired by Skip-Thoughts, as we assume that neighbouring sentences are semantically similar and therefore suitable for training sentence representations using a distributional method. However, instead of learning to predict the individual words in the sentences, we can directly optimise for sentence-level vector similarity.

Given sentence $u$, we randomly pick another nearby sentence $v$ using a normal distribution with a standard deviation of $2.5$. This often gives us neighbouring sentences, but occasionally samples from further away.
We also obtain a negative example $z$ by randomly picking a sentence from the corpus, as this is unlikely to be semantically related to $u$.

Next, each of these sentences is mapped to a vector space by applying the corresponding weights and summing the individual word vectors:

$$
\vec{u} = \sum_{w \in u} g_w \vec{w}
$$

\noindent where $\vec{u}$ is the sentence vector for $u$, $\vec{w}$ is the original embedding for word $w$, and $g_w$ is the learned weight for word $w$.

The following cost function is minimised for training the model -- it optimises the dot product of $u$ and $v$ to have a high value, indicating high vector similarity, while optimising the dot product of $u$ and $z$ to have low values:

$$
cost = max(- \vec{u} \vec{v} + \vec{u} \vec{z}, 0)
$$

\noindent Before the cost calculation, we normalise all the sentence vectors to have unit length, which makes the dot products equivalent to calculating the cosine similarity score. The $max()$ operation is added, in order to stop optimising on sentence pairs that are already sufficiently discriminated. The BNC was used as the text source, and the model was trained with gradient descent and learning rate $0.1$. 

We removed any tokens containing an underscore in the pretrained vectors, as these are used to represent longer phrases, and were left with a vocabulary of $92,902$ words. 
During training, the original word embeddings are left constant, and only the word weights $g_w$ are optimised. This allows us to retrofit the vectors for our specific task with a small number of parameters -- the full embeddings contain $27,870,600$ parameters, whereas we need to optimise only $92,902$.

Similar methods could potentially be used for adapting word embeddings to other tasks, while still leveraging all the information available in the Word2Vec pretrained vectors. We make the trained weights from our system publicly available, as these can be easily used for constructing improved sentence representations for related applications.\footnote{http://www.marekrei.com/projects/weighted-embeddings}

\section{Evaluation}

Since there is no publicly available dataset that contains manually annotated relevance scores at the sentence level, we measure the accuracy of the methods at identifying the original prompt which was used to generate each sentence in a learner essay. While not all sentences in an essay are expected to directly convey the prompt, any noise in the dataset equally disadvantages all systems, and the ability to assign a higher score to the correct prompt directly reflects the ability of the model to capture topical relevance.

Two separate publicly available corpora of learner essays, written by upper-intermediate level language learners, were used for evaluation.
The First Certificate in English dataset (FCE, \newcite{Yannakoudakis2011}), consisting of 30,899 sentences written in response to 60 prompts; and the International Copus of Learner English dataset (ICLE, \newcite{Granger2009}) containing 20,883 sentences, written in response to 13 prompts.\footnote{We used the same ICLE subset as \newcite{Persing2014}.}

There are substantial differences in the types of prompts used in these two datasets. The ICLE prompts are short and general, designed to point the student towards an open discussion around a topic. In contrast, the FCE contains much more detailed prompts, describing a scenario or giving specific instructions on what should be mentioned in the text.
An average prompt in ICLE contains $1.5$ sentences and $19$ words, whereas an average prompt in FCE has $10.3$ sentences and $107$ words. These differences are large enough to essentially create two different variants of the same task, and we will see in Section \ref{sec:results} that alternative methods perform best for each of them.

During evaluation, the system is presented with each sentence independently and aims to correctly identify the prompt that the student was writing to. For longer prompts, the vectors for individual sentences are averaged together. Performance is evaluated through classification accuracy and mean reciprocal rank \cite{Voorhees1999a}.

\section{Results}
\label{sec:results}

\begin{table}
\setlength\tabcolsep{5.0pt}
\begin{tabular}{lrrrr} \toprule
 & \multicolumn{2}{c}{FCE} & \multicolumn{2}{c}{ICLE} \\ 
 & {\footnotesize ACC} & {\footnotesize MRR} & {\footnotesize ACC} & {\footnotesize MRR} \\ \midrule
Random & 1.8 & 7.9 & 7.7 & 24.4 \\
Majority & 22.4 & 25.8 & 28.0 & 39.3 \\
TF-IDF & \textbf{37.2} & \textbf{47.0} & 32.3 & 46.9 \\
Word2Vec & 14.1 & 26.2 & 32.8 & 49.6 \\
IDF-Embeddings & 22.7 & 33.9 & 40.7 & 55.1 \\
Skip-Thoughts & 2.8 & 9.1 & 21.9 & 37.9 \\
Weighted-Embeddings & 24.2 & 35.1 & \textbf{51.5} & \textbf{65.4} \\ \midrule
Combination & 32.6 & 43.4 & 49.8 & 64.1\\ \bottomrule
\end{tabular}
\caption{Accuracy and mean reciprocal rank for the task of sentence-level topic detection on FCE and ICLE datasets.}
\label{tab:results}
\end{table}

\begin{table*}
\small
\setlength\tabcolsep{8pt}
\begin{tabular}{r|l} \toprule
0.382 & Students have to study subjects which are not closely related to the subject they want to specialize in . \\
0.329 & In order for that to happen however , our government has to offer more and more jobs for students . \\
0.085 & I thought the time had stopped and the day on which the results had to be announced never came . \\ \midrule
\multicolumn{2}{l}{University,
degrees,
undergraduate,
doctorate,
professors,
university,
degree,
professor,
PhD,
College,
psychology
} \\
\bottomrule
\end{tabular}
\caption[caption]{Above: Example sentences from essays written in response to the prompt "Most University degrees are theoretical and do not prepare us for the real life. Do you agree or disagree?", and relevance scores using the Weighted-Embeddings method.\\\hspace{\textwidth}Below: Most highly ranked individual words for the same prompt.}
\label{tab:examples}
\end{table*}

Results for all the systems can be seen in Table \ref{tab:results}.
TF-IDF achieves good results and the best performance on the FCE essays. The prompts in this dataset are long and detailed, containing specific keywords and names that are expected to be used in the essay, which is why this method of measuring word overlap achieves the highest accuracy. In contrast, on the ICLE dataset with more general and open-ended prompts, the TF-IDF method achieves mid-level performance and is outranked by several embedding-based methods.

Word2Vec is designed to capture more general word semantics, as opposed to identifying specific tokens, and therefore it achieves better performance on the ICLE dataset. By combining the two methods, in the form of IDF-Embeddings, accuracy is consistently improved on both datasets, confirming the hypothesis that weighting word embeddings can lead to a better sentence representation.

The Skip-Thoughts method does not perform well for the task of sentence-level topic detection. This is possibly due to the model being trained to predict individual words in neighbouring sentences, therefore learning various syntactic and paraphrasing patterns, whereas prompt relevance requires more general topic similarity.
Our results are consistent with those of \newcite{Hill2016a}, who found that Skip-Thoughts performed very well when the vectors were used as features in a separate supervised classifier, but gave low results when used for unsupervised similarity tasks.

The newly proposed Weighted-Embeddings method substantially outperforms Word2Vec and IDF-Embeddings on both datasets, showing that automatically learning word weights in combination with pretrained embeddings is a beneficial approach. In addition, this method achieves the best overall performance on the ICLE dataset by a large margin.

Finally, we experimented with a combination method, creating a weighted average of the scores from TF-IDF and Weighted-Embeddings.
The combination does not outperform the individual systems, demonstrating that these datasets indeed require alternative approaches.
However, it is the second-best performing system on both datasets, making it the most robust method for scenarios where the type of prompt is not known in advance.

\begin{table}[bh]
\small
\setlength\tabcolsep{14.5pt}
\begin{tabular}{rr|rr} \toprule
two & -1.31 & cos & 3.32 \\ 
although & -1.26 & studio & 2.22 \\ 
which & -1.09 & Labour & 2.18 \\ 
five & -1.06 & want & 2.01 \\ 
during & -0.80 & US & 2.00 \\ 
the & -0.73 & Secretary & 1.99 \\ 
unless & -0.66 & Ref & 1.98 \\ 
since & -0.66 & film & 1.98 \\ 
when & -0.66 & v. & 1.91 \\ 
also & -0.65 & Cup & 1.89 \\ \bottomrule
\end{tabular}
\caption{Top lowest and highest ranking words and their weights, as learned by the Weighted-Embeddings method.}
\label{tab:topwords}
\end{table}

\section{Discussion}

In Table \ref{tab:examples} we can see some example learner sentences from the ICLE dataset, together with scores from the Weighted-Embeddings system. The method manages to capture an intuitive relevance assessment for all three sentences, even though none of them contain meaningful keywords from the prompt.
The second sentence receives a slightly lower score compared to the first, as it introduces a somewhat tangential topic of government.
The third sentence is ranked very low, as it contains no information specific to the prompt.
Automated assessment systems relying only on grammatical error detection would likely assign similar scores to all of them.
The method maps sentences into the same vector space as individual words, therefore we are also able to display the most relevant words for each prompt, which could be useful as a writing guide for low-level students.

Table \ref{tab:topwords} contains words with the highest and lowest weights, as assigned by Weighted-Embeddings during training.
We can see that the model has independently learned to disregard common stopwords, such as articles, conjunctions, and particles, as they rarely contribute to the general topic of a sentence.
In contrast, words with the highest weights mostly belong to very well-defined topics, such as politics, entertainment, or sports.

\section{Conclusion}

In this paper, we investigated the task of assessing sentence-level prompt relevance in learner essays. 
Frameworks for evaluating the topic of individual sentences would be useful for capturing unsuitable topic shifts in writing, providing more detailed feedback to the students, and detecting subversion attacks on automated assessment systems.

We found that measuring word overlap, weighted by TF-IDF, is the best option when the writing prompts contain many details that the student is expected to include. However, when the prompts are relatively short and designed to encourage a discussion, which is common in examinations at higher proficiency levels, then measuring vector similarity using word embeddings performs consistently better.

We extended the well-known Word2Vec embeddings by weighting them with IDF, which led to improvements in sentence representations.
Based on this, we constructed the Weighted-Embeddings model for automatically learning individual weights in a data-driven manner, using only plain text as input.
The resulting method consistently outperforms the Word2Vec and IDF-Embeddings methods on both datasets, and substantially outperforms any other method on the ICLE dataset.

\bibliography{references}
\bibliographystyle{naaclhlt2016}

\end{document}